\documentclass{article}

\usepackage{PRIMEarxiv}

\usepackage[utf8]{inputenc} 
\usepackage[T1]{fontenc}    
\usepackage{hyperref}       
\usepackage{url}            
\usepackage{booktabs}       
\usepackage{amsfonts}       
\usepackage{nicefrac}       
\usepackage{microtype}      
\usepackage{lipsum}
\usepackage{fancyhdr}       
\usepackage{graphicx}       
\usepackage{multirow}
\usepackage{bm}
\usepackage{amsmath}
\usepackage{bbm}
\graphicspath{{media/}}     

\title{MP-GELU Bayesian Neural Networks: \\
Moment Propagation by GELU Nonlinearity}

\author{
  Yuki Hirayama, Sinya Takamaeda-Yamazaki \\
  Graduate School of Information Science and Technology \\
  The University of Tokyo\\
  Tokyo, Japan\\
  \texttt{\{hirayama\_yuki, shinya\}@is.s.u-tokyo.ac.jp} \\
}

\begin{document}
\maketitle

\begin{abstract}
Bayesian neural networks (BNNs) have been an important framework in the study of uncertainty quantification.
Deterministic variational inference, one of the inference methods, utilizes moment propagation to compute the predictive distributions and objective functions.
Unfortunately, deriving the moments requires computationally expensive Taylor expansion in nonlinear functions, such as a rectified linear unit (ReLU) or a sigmoid function.
Therefore, a new nonlinear function that realizes faster moment propagation than conventional functions is required.
In this paper, we propose a novel nonlinear function named moment propagating-Gaussian error linear unit (MP-GELU) that enables the fast derivation of first and second moments in BNNs.
MP-GELU enables the analytical computation of moments by applying nonlinearity to the input statistics, thereby reducing the computationally expensive calculations required for nonlinear functions.
In empirical experiments on regression tasks, we observed that the proposed MP-GELU provides higher prediction accuracy and better quality of uncertainty with faster execution than those of ReLU-based BNNs.

\end{abstract}


\section{Introduction}
During the last decade, deep learning has achieved high accuracy in many tasks and plays an important role in a wide range of applications, including image recognition and natural language processing.
However, several studies have reported that deep learning models perform poorly in expressing the uncertainty about out-of-distribution data \cite{Gal2016Uncertainty, kendall2017, 10.5555/3454287.3455541, 42503}.
Making wrong predictions with overconfidence is a critical drawback in applications where prediction reliability should be considered, such as medical diagnosis or image recognition in automated driving \cite{ijcai2017-661}.
Therefore, uncertainty quantification in machine learning is an increasingly important area for the safety of artificial intelligence systems.
In the case of medical diagnosis, if a model can express uncertainty, we can ask for expert judgment when the model says, ``I don't know.''

A Bayesian neural network (BNN) is one of the key frameworks that can handle epistemic and aleatoric uncertainties to mitigate the overconfidence problem.
Aleatoric uncertainty is the uncertainty inherent in data and appears when the data contain some noise, such as coin tosses.
This type of uncertainty can be captured by modeling the noise in the data, as described in Section \ref{sec:method}.
Epistemic uncertainty can be defined as ignorance about the model and appears when the input data are different from the training data \footnote{For example, in the case of inputting an image of a horse to the trained model that classifies dogs and cats.}.
In non-BNNs, we select one model that fits finite training data.
On the other hand, BNNs consider several models that fit the training data by assuming the probability distributions over the models to capture the epistemic uncertainty.

The most well-known inference method of BNN is variational inference (VI) \cite{10.5555/3045118.3045290, Kingma2014, pmlr-v32-rezende14, Gal2016a}.
In VI, we approximate the posterior distribution of the weights by using tractable parametric distributions, such as a Gaussian or a Bernoulli distribution.
The predictive distribution of the test input data is given by Monte Carlo (MC) sampling.
However, this procedure requires forward propagation multiple times, which can be a serious drawback in applications requiring low latency.

Another proposed method, deterministic variational inference (DVI) \cite{DVI}, does not rely on MC sampling but instead performs moment propagation to derive the predictive distribution and objective function.
In DVI, the outputs of intermediate layers are represented as random variables and propagate these first and second moments.
However, it is difficult to calculate these moments analytically for nonlinear functions such as a rectified linear unit (ReLU) or a sigmoid function.
Therefore, DVI requires Taylor expansion, which is computationally expensive.
ReLU is a simple function that rounds up negative values to zero for deterministic inputs.
However, when the input is a random variable, these nonlinear functions are inappropriate for introducing nonlinearity because a computationally expensive approximation is required to derive these moments.

In this paper, we propose a new nonlinear function named moment propagating-Gaussian error linear unit (MP-GELU), a natural extension of GELU \cite{gelu}.
(i) MP-GELU realizes to derive the intermediate moments analytically in the nonlinear function with less computational complexity.
We also show that (ii) MP-GELU naturally introduces the mean and variance of the distribution function into the training, which are the hyperparameters in GELU.
The results of our empirical experiments demonstrate that (iii) the proposed method provides higher prediction accuracy and better quality of uncertainty with faster inference than those of ReLU-based BNNs in regression tasks.

\section{Methods}
\label{sec:method}
In this section, we formulate the objective function and the predictive distribution of BNNs in regression tasks.
Then, we describe the detailed procedures to calculate the objective function and the predictive distribution using moment propagation instead of MC sampling.
In Section \ref{sec:MP-GELU}, we describe MP-GELU nonlinearity, the proposal in this paper.
In the following, let $D$ be the training data, $\bm{x} \in \mathbb{R}^Q$ be the $Q$-dimensional features of the input, and $y \in \mathbb{R}$ be the label of the output.
Let $\bm{w}$ be the weight parameter of the BNN.
We denote an $N \times N$ matrix with diagonal elements $(a_1, a_2, ..., a_N)$ as $\text{diag}(a_1, a_2, ..., a_N)$.

\subsection{Objective function and predictive distribution}
\label{sec:objective}

The goal of the training is to obtain the posterior distribution of weights $p(\bm{w}|D)$; however, it is difficult to solve it analytically.
Therefore VI searches instead for a distribution $q(\bm{w})$ that approximates $p(\bm{w}|D)$ well.
Here, we consider BNNs with a two-dimensional output $\bm h = (h_1, h_2)$ to model the mean and variance of the outputs.
This means that the predictive distribution of the output $y$ conditioned by $\bm h$ is modeled as $p(y | \bm{h}) = N(y | h_1, e^{h_2})$.
In this way, we can capture the heteroscedastic aleatoric uncertainty.

The objective function here is an expected log-likelihood.
We can compute the expected log-likelihood given the data $D$ as follows, as shown in \cite{DVI}:

\begin{equation}
    \mathbb{E}_{\bm{w} \sim q}[\log p(y|\bm{x}, \bm{w})] = \mathbb{E}_{\bm{h} \sim N(\mathbb{E}[h], \Sigma)}[\log p(y |\bm{h})] = - \frac{1}{2}  \{ \log 2 \pi + \mathbb{E}[h_1] + \frac{\Sigma _{1, 1} + (\mathbb{E}[h_1] - \Sigma _{1, 2} - y)^2 }{e^{\mathbb{E}[h_2] - \Sigma _{2, 2} / 2}} \}.
    \label{eq:ll}
\end{equation}

Here, $h$ is the output random variable of the last layer, which follows a multivariate Gaussian distribution.  
$\Sigma$ is the covariance matrix of $\bm{h}$, and $\Sigma _{i, j}$ is the $(i, j)$th element of $\Sigma$.
We optimize the approximated posterior distribution $q(\bm{w})$ by maximizing equation (\ref{eq:ll}).
Like the expected log-likelihood, the predictive distribution $p(y|\bm{x})$ can be computed by marginalizing over $\bm{h}$.

In this paper, we calculate the objective function and the predictive distribution using moment propagation instead of MC sampling.
With MC sampling, the statistics of $\bm{h}$ in equation (\ref{eq:ll}) are estimated from the samples of $\bm{h}$.
On the other hand, with moment propagation, the statistics $\mathbb{E}[h]$ and $\mathbb{V}[h]$ are calculated directly.
For example, the moments of a matrix product of $\bm{h'} = \bm{w}\bm{h}$ are $\mathbb{E}[\bm{h'}] = \mathbb{E}[\bm{w}] \mathbb{E}[\bm{h}]$ and $\text{Cov}[h'_k, h'_l] = \sum _{i, j} \mathbb{E}[h_i, h_j] \text{Cov}[w_{ki}, w_{lj}] + \mathbb{E}[w_{ki}] \text{Cov} [h_i, h_l] \mathbb{E} [w_{lj}]$.
Here, we use the independency between $h$ and $w$.
In the case of a BNN with a dropout layer \cite{Gal2016a, NIPS2015_5666, 10.5555/3305890.3305939, NIPS2017_84ddfb34, Gal2016Bayesian}, the matrix product in a dense layer can be simplified because the weights in the dense layers are not random variables.
In the dropout layer with a dropout rate $p$, we can compute the output random variable as $\bm{h'}=\text{diag}(\epsilon_1, ..., \epsilon_n) \bm{h}$, where $\epsilon _i \sim \text{Bernoulli}(1 - p)$ independently.
Then, we can compute the moments of $\bm{h'}$ in the same way as for dense layers.

As described above, the moments of the output of each layer can be calculated up to the final layer; the problem, however, is the nonlinear function.
As discussed in Section \ref{sec:related_work}, it is computationally inefficient to derive the moments of the output of ReLU, which requires an approximation of the Taylor expansion \cite{DVI}. 
Therefore, instead of using a ReLU function, this paper proposes MP-GELU as a nonlinear function suitable for moment propagation.

\subsection{MP-GELU nonlinearity}
\label{sec:MP-GELU}

GELU \cite{gelu} is an activation function that combines a stochastic regularization method and a nonlinear transformation for non-BNNs. 
Let $h \in \mathbb{R}$ be an input of GELU and $\epsilon \sim \text{Bernoulli} (\Phi (h))$ be a latent random variable.
Here, $\Phi$ is a cumulative distribution function (CDF) of a standard Gaussian distribution.
Note that $h$ is not a random variable here.
The product of $h$ and $\epsilon$ is less likely to be dropped if the value of $h$ is large.
The reason is that, if $\Phi (h)$ approaches 1, $\epsilon$ has a high probability of being 1.
Conversely, if the value of $h$ is small, $h$ is more likely to drop to 0.

The product of $h$ and $\epsilon$ is a random variable; however, in \cite{gelu}, the stochastic behavior is ignored and only the expected value of $h \cdot \epsilon$ is calculated, as in the following equation:

\begin{equation}
  \text{GELU}(h) = \mathbb{E}[h \cdot \epsilon] = h \cdot p(X \leq h) = h \cdot \Phi(h).
\end{equation}

Here, $X$ is a random variable that follows a standard Gaussian distribution.
Thus, GELU introduces nonlinearity by using the CDF of the standard Gaussian distribution.
The choice of the CDF is treated as a hyperparameter in \cite{gelu}.
Note that GELU nonlinearity was originally proposed as an activation function for non-BNNs.
Therefore, the output is assumed to be treated as a definite value.

In this paper, we extended GELU, named MP-GELU, with the idea of using it as an activation function for BNNs, especially in the case of moment propagation.
The major advantages of this proposal are that (i) the parameters of the CDF are determined automatically by the input statistics and (ii) the output moments can be derived analytically after the calculation of the dropout rate.

MP-GELU uses the statistics of the input random variables to calculate the probability that the input is dropped or not.
On the assumption that the random variables in the dense layers follow a Gaussian distribution \footnote{This assumption also requires other nonlinear functions}, the probability that the random variable $h \sim N(h | \mu, \sigma ^2)$ is less than or equal to zero is

\begin{equation}
  p(h \leq 0) = \int_{-\infty}^{0} p(h) dh = \Phi(-\mu / \sigma).
  \label{eq:dropoutrate}
\end{equation}

We use this value as a dropout rate.
Like in GELU, the CDF of the Gaussian distribution is used to determine the dropout rate.
However, the parameters of the CDF, which are treated as hyperparameters in GELU, are obtained as a byproduct of the moment propagation.
A notable feature of MP-GELU is that it does not apply a nonlinear transformation to random variables but rather to the statistics of random variables.

After obtaining the dropout rate in equation (\ref{eq:dropoutrate}), we derive the moments of the MP-GELU output $h' = h \cdot \epsilon$.
Written in vector notation, this is equal to $\bm{h'} = \text{diag}(\epsilon_1, \epsilon_2, ..., \epsilon_n)\bm{h}$, where $n$ is the dimension of input $h$, $h\sim N(h|\mu_i, \sigma_i ^2)$, $\epsilon_i \sim \text{Bernoulli}(1 - \Phi(-\mu_i/ \sigma_i))$.
The moments of $\bm{h'}$ can be obtained using the same procedure as described in Section \ref{sec:objective}. 
Here, we can assume the independence between $\bm{h}$ and $\bm{\epsilon}$.
The reason is that $\bm{\epsilon}$ only depends on the statistics of $\bm{h}$ and not on the observed $\bm{h}$.


In MP-GELU, the dropout rate is obtained by equation (\ref{eq:dropoutrate}).
If we assume that the variance of each input random variable is 1, the dropout rate will be $\Phi(-\mu)$.
This is the same dropout rate derived in GELU.
In GELU, the input is dropped with a probability of $\Phi(-h)$, where $\Phi(-h)$ is the probability that a Gaussian random variable with mean $h$ and variance of 1 falls below 0.
This shows that MP-GELU is a natural extension of GELU.
Furthermore, if we consider the limit of $\sigma \rightarrow 0$ in GELU, this is equal to ReLU, as shown in \cite{gelu}.
The reason is that ReLU only uses the sign of the input to decide whether to drop or not.

\section{Related works}
\label{sec:related_work}
There are several prior studies on moment propagation in BNNs.
Wu et al. \cite{DVI} proposed BNNs that assume a Gaussian distribution for the weights and derived its moments including the ReLU and sigmoid nonlinearity.
In these nonlinear functions, Taylor expansion is required to calculate the full covariance matrix. 
Brach et al. \cite{DBLP:journals/corr/abs-2007-03293} proposed a moment propagation method for BNNs with a dropout layer.
They considered the ReLU nonlinearity and diagonal approximation of the covariance matrix.
In the case of the diagonal approximation, the variance can be calculated analytically instead of ignoring the correlation of activations.

These prior studies mainly focused on how to derive moments in existing layers, such as a dense layer or a ReLU nonlinearity, and did not consider what types of layers are suitable for moment propagation.
However, calculating the moments in these nonlinear functions requires series expansions, which increase the computational complexity.
Therefore, we propose MP-GELU, a nonlinear function suitable for moment propagation.
The amounts of computations required for the MP-GELU and ReLU layers are summarized in Table \ref{tab:comp_comp}.

\begin{table}[h]
  \centering
  \begin{tabular}{ccccc}
  \hline
  Operation                               & MP-GELU (Full)       & ReLU (Full)       & MP-GELU (Diagonal)          & ReLU (Diagonal)          \\ \hline
  Add/Mul                                 &    $O(n^2)$          &    $O(n^2)$            &    $O(n)$          & $O(n)$               \\
  Sqrt/Exp/Erf/Arcsine                     &    $O(n)$            &    $O(n^2)$            &    $O(n)$          &    $O(n)$         \\ \hline
  \end{tabular}
  \caption{Comparison of the number of operations required for MP-GELU and ReLU nonlinearity.
  $n$ is the dimension of the input vector.
  The covariance matrices of activations are computed for all elements in `Full,' whereas `Diagonal' approximates these matrices using diagonal matrices.
  }
  \label{tab:comp_comp}
  \end{table}

In the case of no approximation, MP-GELU requires less computation than ReLU.
In the case of the diagonal approximation, the computational order is the same.
However, as shown in Section \ref{sec:experiments}, MP-GELU is faster than ReLU with a diagonal approximation. 
This is because the roles of the dropout layer and nonlinear functions are realized by a single MP-GELU.

\section{Experiments}
\label{sec:experiments}
In this section, we assess the performance of MP-GELU compared with ReLU in several regression tasks.
In Section \ref{sec:toydata}, we qualitatively evaluate whether two types of uncertainty can be captured by the proposed method on a toy dataset.
In Section \ref{sec:UCIdata}, we quantitatively evaluate the predictive uncertainty, accuracy, and runtime of the two methods on several UCI datasets \cite{Dua:2019}.

Both models compute the expected log-likelihood and predictive distribution deterministically using moment propagation.
The difference between the two models is the choice of the nonlinear function: MP-GELU or ReLU.
MP-GELU-based BNNs use the dropout layer only in the first layer, whereas ReLU-based BNNs use the dropout layer before each dense layer \footnote{The MP-GELU nonlinearity plays the roles of both the dropout layer and nonlinear function; however, in the case where the input is not a random variable, the dropout rate cannot be calculated as in equation \ref{eq:dropoutrate}. Therefore, we use the dropout layer in the first layer to introduce perturbation.}.
The network structures are shown in Tables \ref{tab:MP-gelu} and \ref{tab:relu}.

\begin{table}[h]
  \centering
  \begin{tabular}{cccccccc}
  \hline
  Layer        & Input & Dropout & Dense & MP-GELU & Dense & MP-GELU & Dense \\
  Output shape & Q     & 20      & 20    & 20      & 20    & 20      & 2     \\ \hline
  \end{tabular}
  \caption{Network structure of the MP-GELU-based BNN }
  \label{tab:MP-gelu}
\end{table}

\begin{table}[h]
  \centering
  \begin{tabular}{cccccccccc}
  \hline
  Layer        & Input & Dropout & Dense & ReLU & Dropout & Dense & ReLU & Dropout & Dense \\
  Output shape & Q     & 20      & 20    & 20   & 20      & 20    & 20   & 20      & 2    \\ \hline
  \end{tabular}
  \caption{Network structure of the ReLU-based BNN}
  \label{tab:relu}
\end{table}

The output of the network has two units because the mean and variance of the predictive distribution are modeled at the same time, as explained in Section \ref{sec:method}.
We use a dropout layer in the first layer to introduce perturbation into MP-GELU-based BNN.
Another option is to insert GELU (not MP) into the first layer; however, we find that it degrades the predictive performance.
Therefore, we introduce perturbations using the dropout layer in the experiments.

We can apply MP-GELU as a nonlinear function regardless of the approximate posterior distribution of the weights.
However, in this paper, we chose a BNN with a dropout layer for the experiments for the following reasons: (i) it is easy to compute the covariance matrix in the dense layer because the weight parameters are not random variables in BNNs with a dropout layer, (ii) it has been reported that BNNs with a dropout layer perform better than BNNs with a Gaussian distribution \cite{nado2021uncertainty}, and (iii) it prevents the increase of hyperparameters for the training \footnote{For example, the initializer of the mean and variance for the weights.}.

\subsection{Toy dataset}
\label{sec:toydata}

In this section, we qualitatively assess whether MP-GELU and ReLU can capture the epistemic and aleatoric uncertainty on a toy dataset.
We generated the toy datasets using the following equations:
\begin{gather*}
  y_i = \sin (2 x_i) \cdot \cos(7 x_i) + \epsilon _i \\
  \epsilon_i \sim N(\epsilon_i | 0, (\sin x_i)^2) \\
  x_i \sim U(-0.5, 0.5)
\end{gather*}
To generate data with a heteroscedastic aleatoric uncertainty, we added $\epsilon_i$ as noise, which depends on the input $x_i$.
The training data were sampled from a uniform distribution $U[-0.5, 0.5]$, and the test data were in the range of $[-1, 1]$ to intentionally generate an out-of-distribution dataset.
There were 100 data in the training set.

In the training, we used stochastic gradient descent (SGD) as an optimizer.
The learning rate was set to 0.1, and the training epoch was 1000.
The batch size was 100 for both methods.
The dropout rate was set to 0.001 for both methods.
In both methods, the covariance matrix is fully computed without a diagonal approximation.

\begin{figure}[h]
    \centering
  \begin{minipage}[b]{0.49\linewidth}
    \includegraphics[scale=0.53]{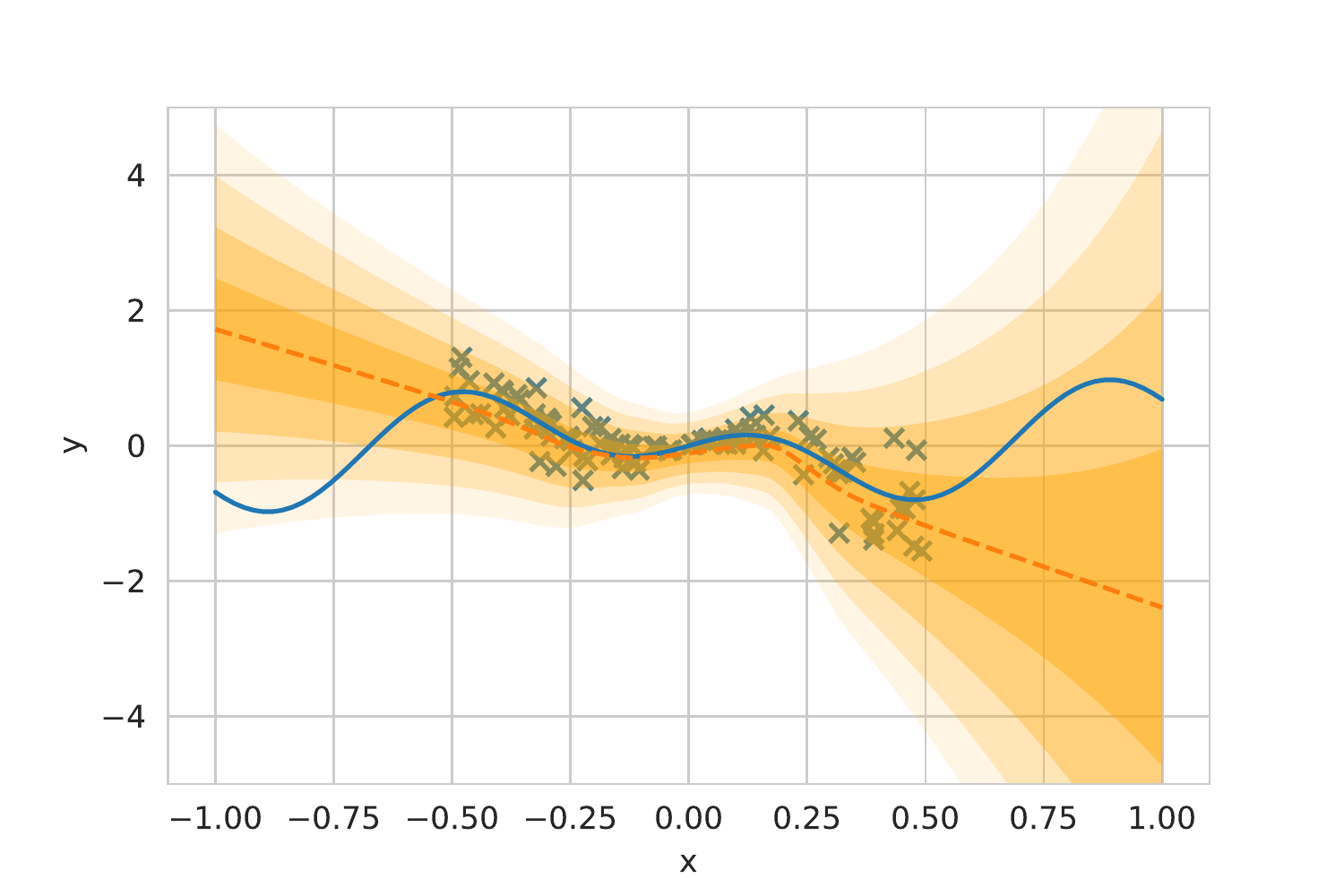}
  \end{minipage}
  \begin{minipage}[b]{0.49\linewidth}
    \includegraphics[scale=0.53]{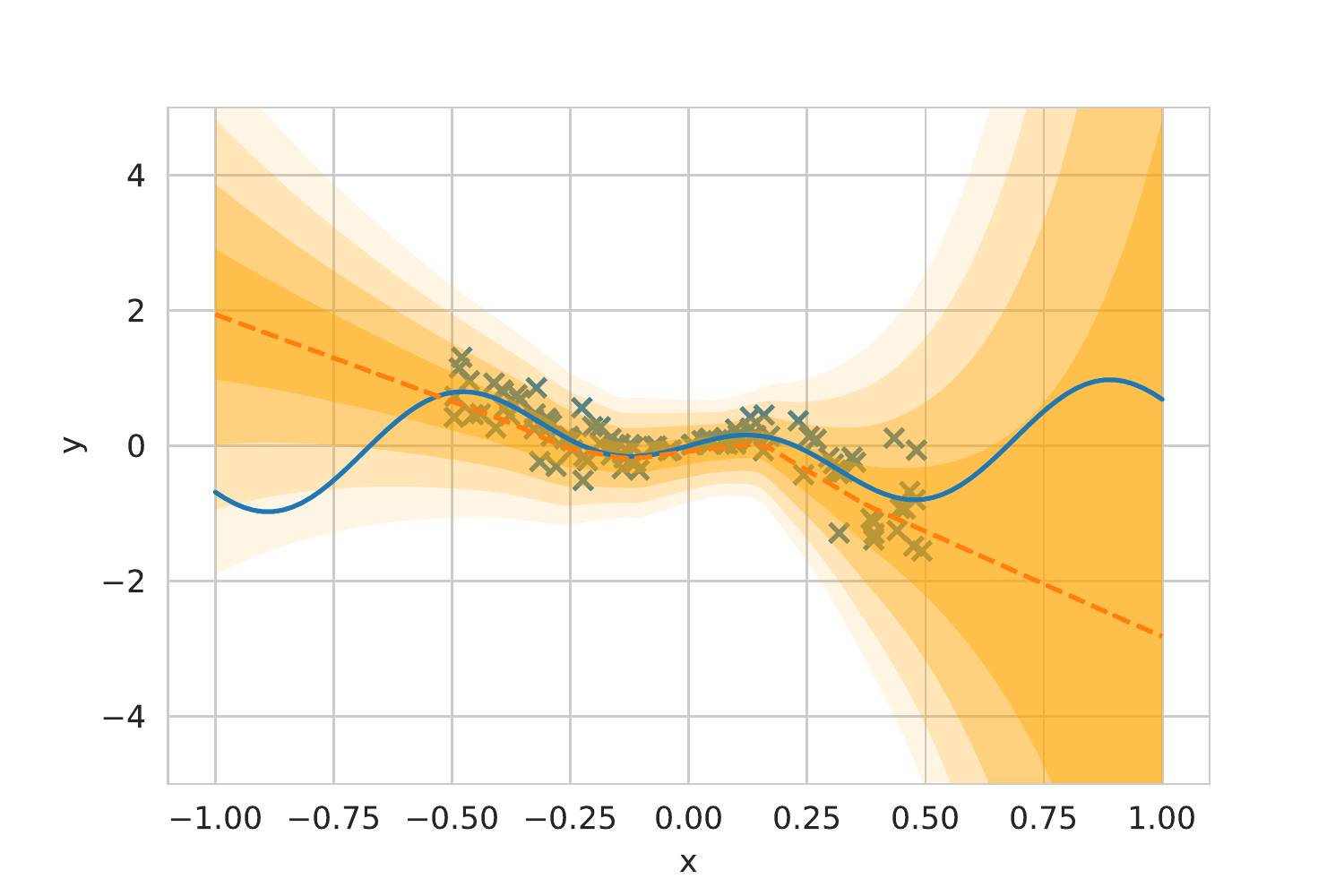}
  \end{minipage}
  \caption{Predictive distributions of the ReLU-based BNN (left) and the MP-GELU-based BNN (right).
  The blue crosses denote the training data, and the blue curve is the true function that generates the training data.
  The orange dashed line is the mean of the prediction, and the orange-filled region shows the standard deviations ($\sigma, 2\sigma, ..., 4\sigma$) of the predictive distribution.
  }
  \label{fig:toydata}
\end{figure}

Figure \ref{fig:toydata} shows the predictive distribution of a ReLU-based BNN and an MP-GELU-based BNN.
As shown in Figure \ref{fig:toydata}, both methods have a higher variance in regions where there are no training data, such as in $[-1, -0.5]$ or $[0.5, 1]$.
This indicates that they can capture the epistemic uncertainty.
In a region $[-0.5, 0.5]$ where the training data exist, we can observe that these methods capture the data-dependent noise.
The above qualitative results on the toy dataset show that the MP-GELU-based BNN can capture the two types of uncertainty as efficiently as the ReLU-based BNN.

\subsection{UCI datasets}
\label{sec:UCIdata}
In this section, we assess the MP-GELU-based and ReLU-based BNNs on several UCI datasets \cite{Dua:2019}.
The network structures are the same as in Section \ref{sec:toydata}.
We provide the detailed settings, such as the hyperparameters, dataset partitioning, and preprocessing, in Appendix \ref{sec:uci_setting}.
The evaluation metrics are the negative log-likelihood (NLL), root mean squared error (RMSE), and runtime.
For these metrics, the lower the better.
Runtime is the required time to perform inference on the test data.

\begin{table}[ht]
\centering
\small
\begin{tabular}{ccccccccc}
\hline
\multirow{2}{*}{} & \multirow{2}{*}{$N$} & \multirow{2}{*}{$Q$} & \multicolumn{2}{c}{Test NLL} & \multicolumn{2}{c}{Test RMSE} & \multicolumn{2}{c}{Test runtime [s]} \\
 &  &  & MP-GELU & ReLU & MP-GELU & ReLU & MP-GELU & ReLU \\ \hline
Boston    & 506 & 13 & 1.204 $\pm$ 0.029 & 1.25 $\pm$ 0.027 & 0.786 $\pm$ 0.04 & 0.809 $\pm$ 0.039 & 0.016 & 0.021 \\
Concrete  & 1,030 & 8 & 1.118 $\pm$ 0.022& 1.137 $\pm$ 0.021 & 0.742 $\pm$ 0.016 & 0.748 $\pm$ 0.016 & 0.017 & 0.021 \\
Energy    & 768 & 8 & 0.621 $\pm$ 0.049 & 0.649 $\pm$ 0.047 & 0.46 $\pm$ 0.023 & 0.464 $\pm$ 0.023 & 0.017 & 0022 \\
Kin8nm    & 8,192 & 8 & 0.785 $\pm$ 0.014 & 0.815 $\pm$ 0.013 & 0.561 $\pm$ 0.007 & 0.567 $\pm$ 0.007 & 0.039 & 0.058 \\
Naval     & 11,934 & 16 & 1.21 $\pm$ 0.02 & 1.228 $\pm$ 0.017 & 0.785 $\pm$ 0.022 & 0.794 $\pm$ 0.019 & 0.048 & 0.071 \\
Power     & 9,568 & 4 & 0.035 $\pm$ 0.07 & 0.06 $\pm$ 0.006  & 0.247 $\pm$ 0.002 & 0.246 $\pm$ 0.002 & 0.04 & 0.06 \\
Protein   & 45,730 & 9 & 1.123 $\pm$ 0.01& 1.131 $\pm$ 0.007 & 0.789 $\pm$ 0.003 & 0.793 $\pm$ 0.003 & 0.146 & 0.232 \\
Wine      & 1,599 & 11 & 1.249 $\pm$ 0.016 & 1.252 $\pm$ 0.016 & 0.848 $\pm 0.013 $& 0.849 $\pm$ 0.013 & 0.017 & 0.022 \\
Yacht     & 308 & 6 & 1.307 $\pm$ 0.046 & 1.333 $\pm$ 0.043& 0.879 $\pm$ 0.045 &0.892 $\pm$ 0.044 & 0.016 & 0.021 \\ \hline
\end{tabular}
\caption{
Performance of the MP-GELU-based and ReLU-based BNNs.
$N$ is the data size and $Q$ is the dimension of the features of each dataset.
We show the mean and standard error for each indicator.
The standard error of the test runtime is omitted because of its smallness.
}
\label{tab:uci1}
\end{table}

Table \ref{tab:uci1} shows the results for each dataset.
As shown in the table, MP-GELU performs better than ReLU in terms of both NLL and RMSE in almost all datasets.
A possible explanation for these results is that, as noted in \cite{gelu}, there is a curvature for positive values in GELU, which may allow it to approximate more complex functions than those of ReLU.
The runtimes for MP-GELU are, on average, $27\% $ faster than those of ReLU.
This is because MP-GELU is less computationally expensive than ReLU, as shown in Table \ref{tab:comp_comp}.
In Appendix \ref{sec:appendix_uci}, we show the results of the other settings, e.g., prediction with a single output, and approximate the covariance matrix using a diagonal matrix.
In these settings, we also observed that MP-GELU-based BNNs perform better than ReLU-based BNNs.

\section{Conclusion}
We have presented MP-GELU, a nonlinear function that enables the straightforward derivation of moments in BNNs.
Whereas previous studies focused on how to derive the moments of existing layers, this paper focused on the ease of moment propagation and showed that moments can be derived analytically by extending GELU.
MP-GELU enables the analytical computation of moments by applying nonlinear functions to input statistics, thereby reducing the computationally expensive functions required for nonlinear functions.

In the empirical experiments, we observed that the proposed MP-GELU-based BNN achieves higher prediction accuracy and better quality of uncertainty than those of the BNN with ReLU nonlinearity.
Furthermore, the proposed method was able to perform up to $27\%$ faster inference than that of ReLU-based BNNs.

\section*{Acknowledgments}
This work was supported by JST PRESTO JPMJPR18M9, JST CREST JPMJCR21D2, and JSPS KAKENHI grant no. JP21J20357.

\bibliographystyle{unsrt}
\bibliography{references}

\newpage
\section{Appendix}
\subsection{Settings of the experiments on the UCI dataset}
\label{sec:uci_setting}
We used an SGD optimizer in all the datasets and in both methods.
The learning rate was 0.001, and the number of epochs was 500.
The batch size was 256.

We divided the dataset into the training and test sets as follows.
We randomly selected 10\% of the data as the test set and treated the remaining data as the training set.
We repeated this procedure 20 times to generate 20 sets of training and test sets.

We decided the dropout rate by performing a grid search using the following procedure for both MP-GELU and ReLU.
At first, we split the data into the training and test set and 20\% of the training set was used as the validation set.
After training the model on the remaining 80\%, we measured the NLL on the validation set.
We repeated this procedure for 20 sets of datasets and calculated the average NLL.
We measured the average NLL in the same way as for the other dropout rates, and we selected the dropout rate with the lowest NLL.
The search space was $[0.005, 0.01, 0.05, 0.1]$.
The grid search schemes and the data partitioning were the same as those in \cite{Gal2016a}.

The input features of the training data were normalized so that each feature had a mean of 0 and a variance of 1 across the data.
The labels of the training set were also normalized in the same way; however, the test set was normalized using the statistics of the training set.

\subsection{Evaluation on the UCI datasets}
\label{sec:appendix_uci}

In this section, we present the results of the evaluation on the UCI dataset in other settings.
Table \ref{tab:ou2diag} shows the results of approximating the activation covariance using a diagonal matrix.
Table \ref{tab:ou1full} shows the results when the model has one output unit.
In this model, the predictive distribution is $p(y|x, w) = \mathbbm{1}_{h_1}$.
Table \ref{tab:ou1diag} is the diagonal approximation of this model.

\begin{table}[h]
  \centering
  \small
  \begin{tabular}{ccccccccc}
  \hline
  \multirow{2}{*}{} & \multirow{2}{*}{$N$} & \multirow{2}{*}{$Q$} & \multicolumn{2}{c}{Test NLL} & \multicolumn{2}{c}{Test RMSE} & \multicolumn{2}{c}{Test runtime [s]} \\
   &  &  & MP-GELU & ReLU & MP-GELU & ReLU & MP-GELU & ReLU \\ \hline
  Boston    & 506 & 13 &  1.208 $\pm$ 0.029 & 1.255 $\pm$ 0.027   &  0.785 $\pm$ 0.04 & 0.809 $\pm$ 0.039 & 0.014 & 0.015 \\
  Concrete  & 1,030 & 8 & 1.118 $\pm$  0.022  & 1.136  $\pm$ 0.021 & 0.742 $\pm$ 0.016 & 0.748 $\pm$ 0.016 & 0.015 & 0.016 \\
  Energy    & 768 & 8 & 0.618 $\pm$ 0.05  & 0.647 $\pm$ 0.047 & 0.463 $\pm$ 0.023 &  0.46 $\pm$  0.023 & 0.014 & 0.016 \\
  Kin8nm    & 8,192 & 8 & 0.791 $\pm$ 0.014 &  0.805 $\pm$ 0.014 & 0.563 $\pm$ 0.007 &  0.567 $\pm$  0.007 & 0.033 & 0.036 \\
  Naval     & 11,934 & 16 & 0.997 $\pm$ 0.056 & 1.009 $\pm$ 0.05 & 0.693 $\pm$ 0.037 &  0.687 $\pm$  0.035 & 0.038 & 0.043 \\
  Power     & 9,568 & 4 & 0.02 $\pm$ 0.008 & 0.039 $\pm$ 0.007 & 0.246 $\pm$ 0.002 & 0.246 $\pm$ 0.002 & 0.033  & 0.035 \\
  Protein   & 45,730 & 9 & 1.12 $\pm$ 0.009 & 1.124 $\pm$ 0.008 & 0.79 $\pm$ 0.003 & 0.792 $\pm$ 0.003 & 0.113  & 0.13 \\
  Wine      & 1,599 & 11 & 1.249 $\pm$ 0.016 & 1.252 $\pm$ 0.016 & 0.848 $\pm$ 0.013 & 0.849 $\pm$ 0.013 & 0.014  & 0.015 \\
  Yacht     & 308 & 6 & 1.306 $\pm$ 0.046 & 1.336 $\pm$ 0.043 & 0.878 $\pm$ 0.045 & 0.892 $\pm$ 0.044 & 0.015  & 0.015 \\ \hline
  \end{tabular}
  \caption{Output unit: 2; covariance: diagonal}
  \label{tab:ou2diag}
  \end{table}

  \begin{table}[h]
  \centering
  \small
  \begin{tabular}{ccccccccc}
  \hline
  \multirow{2}{*}{} & \multirow{2}{*}{$N$} & \multirow{2}{*}{$Q$} & \multicolumn{2}{c}{Test NLL} & \multicolumn{2}{c}{Test RMSE} & \multicolumn{2}{c}{Test runtime [s]} \\
   &  &  & MP-GELU & ReLU & MP-GELU & ReLU & MP-GELU & ReLU \\ \hline
  Boston    & 506 & 13 & 1.239 $\pm$ 0.024 & 1.285 $\pm$ 0.024 & 0.767 $\pm$ 0.03 & 0.81 $\pm$ 0.029 & 0.016 & 0.021 \\
  Concrete  & 1,030 & 8 & 1.193 $\pm$ 0.009 & 1.197 $\pm$ 0.009 & 0.737 $\pm$ 0.013 & 0.74 $\pm$ 0.013 & 0.015 & 0.02 \\
  Energy    & 768 & 8 & 1.054 $\pm$ 0.011 & 1.057 $\pm$ 0.011 & 0.511 $\pm$ 0.02 & 0.513 $\pm$ 0.02 & 0.016 & 0.021 \\
  Kin8nm    & 8,192 & 8 & 1.12 $\pm$ 0.005 & 1.122 $\pm$ 0.005 & 0.627 $\pm$ 0.008 & 0.629 $\pm$ 0.008 & 0.035 & 0.056 \\
  Naval     & 11,934 & 16 & 1.311 $\pm$ 0.009 & 1.296 $\pm$ 0.011 & 0.854 $\pm$ 0.014 & 0.835 $\pm$ 0.017 & 0.041 & 0.067 \\
  Power     & 9,568 & 4 & 0.955 $\pm$ 0     & 0.957 $\pm$ 0     & 0.262 $\pm$ 0.002 & 0.262 $\pm$ 0.002 &  0.035 & 0.057 \\
  Protein   & 45,730 & 9 & 1.24 $\pm$ 0.002 & 1.244 $\pm$ 0.001 & 0.796 $\pm$ 0.002 & 0.797 $\pm$ 0.001 & 0.123 & 0.205 \\
  Wine      & 1,599 & 11 & 1.279 $\pm$ 0.011 & 1.281 $\pm$ 0.011 & 0.846 $\pm$ 0.012 & 0.847 $\pm$ 0.013 & 0.016 & 0.016 \\
  Yacht     & 308 & 6 &  1.324 $\pm$ 0.039 & 1.355 $\pm$ 0.039 & 0.866 $\pm$ 0.046 & 0.894 $\pm$ 0.045 & 0.015 & 0.021 \\ \hline
  \end{tabular}
  \caption{Output unit: 1; covariance: full}
  \label{tab:ou1full}
  \end{table}
  
  \begin{table}[h]
  \centering
  \small
  \begin{tabular}{ccccccccc}
  \hline
  \multirow{2}{*}{} & \multirow{2}{*}{$N$} & \multirow{2}{*}{$Q$} & \multicolumn{2}{c}{Test NLL} & \multicolumn{2}{c}{Test RMSE} & \multicolumn{2}{c}{Test runtime [s]} \\
   &  &  & MP-GELU & ReLU & MP-GELU & ReLU & MP-GELU & ReLU \\ \hline
  Boston    & 506 & 13 & 1.241 $\pm$ 0.023 & 1.287 $\pm$ 0.024 & 0.766 $\pm$ 0.03 & 0.809 $\pm$ 0.029 & 0.012 & 0.014 \\
  Concrete  & 1,030 & 8 & 1.193 $\pm$ 0.009 & 1.197 $\pm$ 0.009 & 0.737 $\pm$ 0.013 & 0.74 $\pm$ 0.013 & 0.014 & 0.015 \\
  Energy    & 768 & 8 & 1.054 $\pm$ 0.011 & 1.057 $\pm$ 0.011 & 0.51 $\pm$ 0.02 & 0.513 $\pm$ 0.02 & 0.013 & 0.014 \\
  Kin8nm    & 8,192 & 8 & 1.119 $\pm$ 0.005 & 1.121 $\pm$ 0.005 & 0.627 $\pm$ 0.008 & 0.628 $\pm$ 0.008 & 0.029 & 0.032 \\
  Naval     & 11,934 & 16 & 1.247 $\pm$ 0.018 & 1.266 $\pm$ 0.017 & 0.786 $\pm$ 0.025 & 0.808 $\pm$ 0.023 & 0.034 & 0.037 \\
  Power     & 9,568 & 4 & 0.954 $\pm$ 0     & 0.956 $\pm$ 0 & 0.262 $\pm$ 0.002 & 0.262 $\pm$ 0.002 & 0.027 & 0.032 \\
  Protein   & 45,730 & 9 & 1.238 $\pm$ 0.001 & 1.24 $\pm$ 0.001 & 0.796 $\pm$ 0.002 & 0.796 $\pm$ 0.001 & 0.094 & 0.116 \\
  Wine      & 1,599 & 11 & 1.28 $\pm$ 0.011 & 1.281 $\pm$ 0.011 & 0.846 $\pm$ 0.012 & 0.847 $\pm$ 0.013 & 0.014 & 0.014 \\
  Yacht     & 308 & 6 &  1.326 $\pm$ 0.039 & 1.356 $\pm$ 0.039 & 0.865 $\pm$ 0.046 & 0.894 $\pm$ 0.045 & 0.012 & 0.015 \\ \hline
  \end{tabular}
  \caption{Output unit: 1; covariance: diagonal}
  \label{tab:ou1diag}
  \end{table}

\end{document}